\documentclass[11pt,a4paper]{article}
\usepackage[T1]{fontenc}
\usepackage{newtxtext}
\usepackage[a4paper,left=25mm,right=25mm,top=24mm,bottom=25mm,headheight=14pt,headsep=10mm]{geometry}
\usepackage[round,authoryear]{natbib}
\usepackage{titlesec,fancyhdr,caption,tcolorbox}
\usepackage{amsmath,amssymb,booktabs,tabularx,array}
\usepackage{graphicx,microtype}
\usepackage{fontawesome5}
\usepackage[table]{xcolor}
\usepackage{colortbl}
\usepackage{algorithm,algpseudocode}
\usepackage{hyperref,url}
\usepackage{listings,needspace}
\tcbuselibrary{breakable}

\title{Learning to Compose: Recursive Self-Improvement of Shared Executable Stacks for Zero-Shot Long-Horizon Robot Manipulation}
\author{Yangzhi Chen\textsuperscript{1}\quad Feiran Liu\textsuperscript{1}\quad Zishuo Wang\textsuperscript{1}\quad Zijian Jin\textsuperscript{2}\quad Xin Zhu\textsuperscript{3}\quad Lecheng Ruan\textsuperscript{4}\quad Wenjuan Han\textsuperscript{\textcolor{paperblue}{\faEnvelope}}\thanks{Wenjuan Han is the corresponding author.}\textsuperscript{,5}}
\newcommand{\paperaffiliations}{%
\textcolor{paperblue}{\faGraduationCap}\enspace\textsuperscript{1}Beijing Jiaotong University\quad
\textcolor{paperblue}{\faGraduationCap}\enspace\textsuperscript{2}New York University\quad
\textcolor{paperblue}{\faGraduationCap}\enspace\textsuperscript{3}University of Southern California\quad
\textcolor{paperblue}{\faGraduationCap}\enspace\textsuperscript{4}PKU-HEC joint lab for industrial intelligence, Peking University\quad
\textcolor{paperblue}{$\blacklozenge$}\enspace\textsuperscript{5}LingXing Tech}

\newcommand{\papergithub}{https://github.com/cocacola-lab/Recursia}
\newcommand{\paperwebsite}{https://cocacola-lab.github.io/Recursia-website/}
\newcommand{\stack}{\mathcal{S}}
\newcommand{\recursia}{\textsc{Recursia}}

\newcolumntype{Y}{>{\raggedright\arraybackslash}X}
\hypersetup{colorlinks=true,linkcolor=blue!45!black,citecolor=blue!45!black,urlcolor=blue!45!black}

\definecolor{paperblue}{HTML}{174B7A}
\definecolor{linkblue}{HTML}{2463A0}
\definecolor{paleblue}{HTML}{F0F6FC}
\definecolor{headerblue}{HTML}{60758B}
\hypersetup{colorlinks=true,linkcolor=paperblue,citecolor=linkblue,urlcolor=linkblue}
\titleformat{\section}{\large\bfseries\color{black}}{\thesection}{0.65em}{}
\titleformat{\subsection}{\normalsize\bfseries\color{black}}{\thesubsection}{0.65em}{}
\titleformat{\subsubsection}{\normalsize\bfseries}{\thesubsubsection}{0.65em}{}
\titleformat{\paragraph}[runin]{\normalsize\bfseries}{}{0pt}{}
\titlespacing*{\section}{0pt}{18pt plus 3pt minus 2pt}{8pt}
\titlespacing*{\subsection}{0pt}{13pt plus 2pt minus 2pt}{5pt}
\titlespacing*{\paragraph}{0pt}{8pt}{0.5em}
\renewcommand{\headrulewidth}{0pt}
\renewcommand{\footrulewidth}{0pt}
\fancypagestyle{plain}{\fancyhf{}\fancyfoot[C]{\small\color{headerblue}\thepage}\renewcommand{\headrulewidth}{0pt}}
\makeatletter
\renewcommand{\maketitle}{%
  \thispagestyle{plain}%
  \begin{center}
  {\fontsize{26}{30}\selectfont\bfseries\color{paperblue}\raisebox{-4mm}{\includegraphics[height=15mm,trim=230 210 250 190,clip]{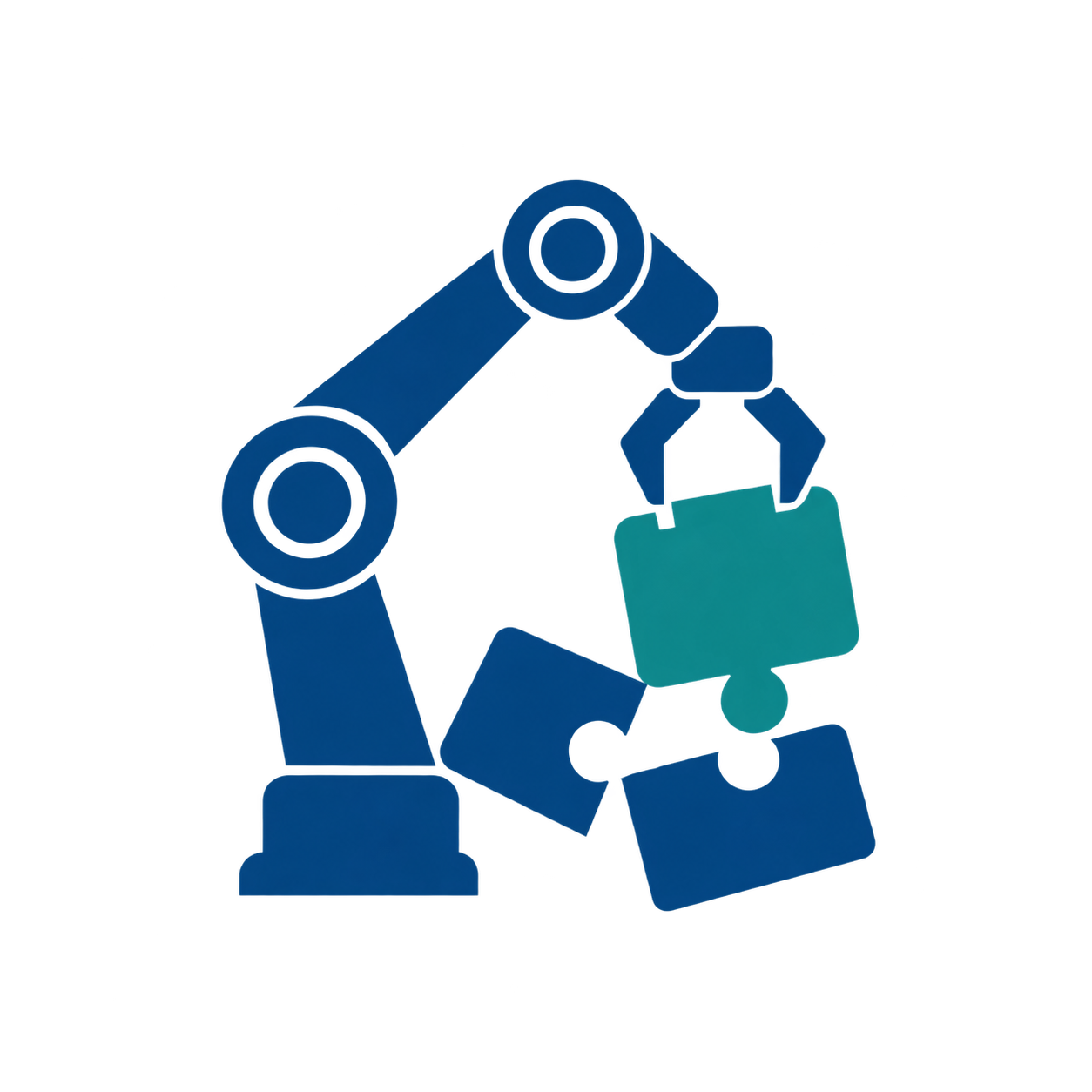}}\hspace{0.3em}Learning to Compose\par}
  \vspace{8pt}
  {\fontsize{16}{20}\selectfont\bfseries
  Recursive Self-Improvement of Shared Executable Stacks\\
  for Zero-Shot Long-Horizon Robot Manipulation\par}
  \vspace{14pt}
  {\normalsize\bfseries\@author\par}
  \vspace{6pt}
  {\small\color{headerblue}\paperaffiliations\par}
  \vspace{10pt}
  {\small\sffamily
  \href{\papergithub}{GitHub: \papergithub}\par
  \vspace{3pt}
  \href{\paperwebsite}{Project Page: \paperwebsite}\par
  \vspace{3pt}
  \href{mailto:wjhan@bjtu.edu.cn}{Email: wjhan@bjtu.edu.cn}\par}
  \vspace{12pt}
  {\color{paperblue}\rule{0.16\linewidth}{1.5pt}\par}
  \end{center}
  \vspace{5pt}}
\makeatother
\renewenvironment{abstract}{%
  \begin{tcolorbox}[colback=black!3,colframe=black,boxrule=0pt,
    leftrule=2pt,arc=0pt,left=12pt,right=12pt,top=10pt,bottom=10pt]
  {\small\sffamily\bfseries\color{black}ABSTRACT\par}\vspace{5pt}
  \small\setlength{\parindent}{0pt}%
}{\end{tcolorbox}\vspace{5pt}}

\lstdefinestyle{appendixprompt}{basicstyle=\fontsize{8.4}{10.4}\selectfont\ttfamily,
  breaklines=true,breakatwhitespace=true,columns=fullflexible,keepspaces=true,
  showstringspaces=false,showspaces=false,showtabs=false,tabsize=2,
  aboveskip=3pt,belowskip=3pt,linewidth=\linewidth}
\newtcolorbox{promptbox}[1]{breakable,
  colback=paleblue!45!white,colframe=paperblue!65!white,
  colbacktitle=paperblue,coltitle=white,fonttitle=\small\bfseries,
  title={#1},title after break={#1 (continued)},boxrule=0.5pt,arc=1.5pt,left=8pt,right=8pt,top=6pt,bottom=6pt}

\date{}

\begin{document}
\maketitle
\begin{abstract}
Individually competent robot skills can fail when composed: tools may lack a required capability, completion signals may misrepresent physical progress, and one skill may leave a state that violates its successor's prerequisites. We introduce \recursia{}, a recursive self-improvement framework that treats the shared executable stack, rather than individual task programs, as the persistent unit of improvement. The stack integrates tools, skills with execution contracts, handoffs, and verifiers. Source-side execution traces guide failure attribution and the generation of missing tools or repairs to existing shared components. Generality checks, mechanism tests, and paired selection and regression gates determine which changes become the next parent. The resulting stack is frozen and reused through declarative task adapters, allowing source-derived changes to support new compositions without target-side code repair. Experiments cover LIBERO-Pro Long simulation and a physical Franka FR3 equipped with a Franka Hand. In simulation, \recursia{} improves success over ASPIRE by 5.8 percentage points under \emph{Pos} perturbations and 15.1 points under \emph{Task} perturbations. Across three physical compositional target tasks, it improves mean success over the Skill Document Only baseline by 40 percentage points. Analyses of constituent execution, candidate rejection, and inference drivers distinguish local mechanism improvements from complete task success and characterize the cost of feedback-conditioned orchestration. These results examine shared executable interfaces as a concrete mechanism for compositional transfer in robot programming.
\end{abstract}
\section{Introduction}
\label{sec:introduction}
Long-horizon robot manipulation demands more than a repertoire of individually capable skills~\citep{mees2022calvin}. Executing a pick-and-place action, for instance, is insufficient if it leaves the environment in a state that violates the preconditions of the subsequent skill~\citep{xu2026baton}. Grounding high-level language plans in skill affordances and perceptual geometry is therefore critical for physically feasible execution~\citep{ahn2022saycan,huang2023voxposer}. 
The central question is how skills acquired on source tasks can be composed to reliably solve previously unseen long-horizon target tasks.

Coding agents provide a natural testbed for studying this compositional generalization~\citep{liang2023code,mu2024robocodex,capx2026}. Language-model-generated programs can invoke perception and control APIs, implement feedback loops, and reuse procedural abstractions~\citep{liang2023code}. RoboCodeX further decomposes multimodal instructions into object-centric manipulation units and synthesizes control code~\citep{mu2024robocodex}, while ProgPrompt incorporates state-feedback assertions and recovery actions into generated plans~\citep{singh2023progprompt}. Skill-library systems such as Voyager store executable procedures and use environmental feedback for extension~\citep{voyager2023}; ASPIRE combines execution feedback, program refinement, skill discovery, and evolutionary search for manipulation~\citep{lu2026aspire}. Lifelong Robot Library Learning expands skill repertoires from accumulated experience~\citep{tziafas2024lrll}, and Uni-Skill requests new skill implementations when existing ones fall short~\citep{uniskill2026}. CaP-X examines how abstraction levels, perceptual grounding, and structured execution feedback affect the reliability of embodied coding agents~\citep{capx2026}. Although these capabilities make program improvement a promising route to transfer, reliable composition depends not only on the generated task program, but critically on the \emph{shared interfaces} it invokes: tools, skill contracts, inter-skill state transfers, and verification logic. Thus, the central object of study shifts from \emph{how} an agent writes a program to \emph{which reusable execution semantics} that program inherits.

We identify four key obstacles to reliable composition: (1) \textbf{Tool insufficiency}, where a task requires a reusable capability absent from the initial API; (2) \textbf{Semantic misalignment}, where an existing tool returns normally despite failing to reach its requested pose of the next tool in the skill; (3) \textbf{State incompatibility}, where a skill's terminal state violates its successor's preconditions~\citep{xu2026baton}; and (4) \textbf{Verification brittleness}, where incorrect checks advance an unfinished task or disturb a completed goal. These failure modes correspond to different interfaces in a hierarchical controller: tool execution, skill completion, inter-skill state transfer, and verifiers.
Prior systems acquire and reuse executable skills from experience~\citep{voyager2023,tziafas2024lrll,lu2026aspire}, and execution-guided self-debugging demonstrates that failed programs yield useful repair evidence~\citep{chen2024selfdebug}. The Darwin G\"odel Machine extends this to persistent self-modification evaluated on coding benchmarks~\citep{zhang2025dgm}. We investigate how repairs to shared interfaces with multiple layers can be validated and retained across tasks. We hypothesize that a repair at a shared interface should improve multiple compositions, whereas a task-specific patch should benefit only the trajectory in which it is written. We examine this hypothesis through simulation and physical-robot experiments on long-horizon compositions.

We propose \recursia{}, a framework that targets compositional generalization by treating the \emph{shared executable stack} as the primary object of persistent improvement. A coding agent can add missing tools, repair skill contracts, or improve handoffs and verifiers. Each modification becomes part of a versioned implementation reused by subsequent source tasks and frozen during target-task execution. 
Specifically, \recursia{} is stack-aware and validation-gated. Source-side execution traces identify the first violated requirement, attributing failures to specific repair layers. This attribution proposes a mechanism for change but does not itself establish efficacy. Candidates must satisfy restrictions on task-specific specialization, survive paired comparisons with the parent stack, and pass the validation gate with selection gate and regression gate. Accepted changes are retained with their failure class, validation evidence, and code version. Task variation is expressed through declarative adapters. At deployment, the frozen stack relies solely on permitted robot observations; evaluator-side ground truth used during source diagnosis is never an input to the deployed policy.

Our contributions are threefold: 
(1) We formulate skill composition as the persistent improvement of a \emph{shared executable stack} integrating tools, skills with execution contracts, handoffs, and verifiers, and enable the generation of previously absent tools. 
(2) We introduce a failure-attributed improvement procedure that localizes repairs to specific stack layers and validates shared patches through generality and regression gates. 
(3) We evaluate frozen-stack transfer to long-horizon compositions in LIBERO-Pro Long simulation and on a physical Franka FR3 robot, comparing with skill-library and skill-document baselines and analyzing compositional execution, validation decisions, and inference strategies.

\begin{figure}[t]
\centering
\includegraphics[width=0.97\linewidth,height=0.74\textheight,keepaspectratio]{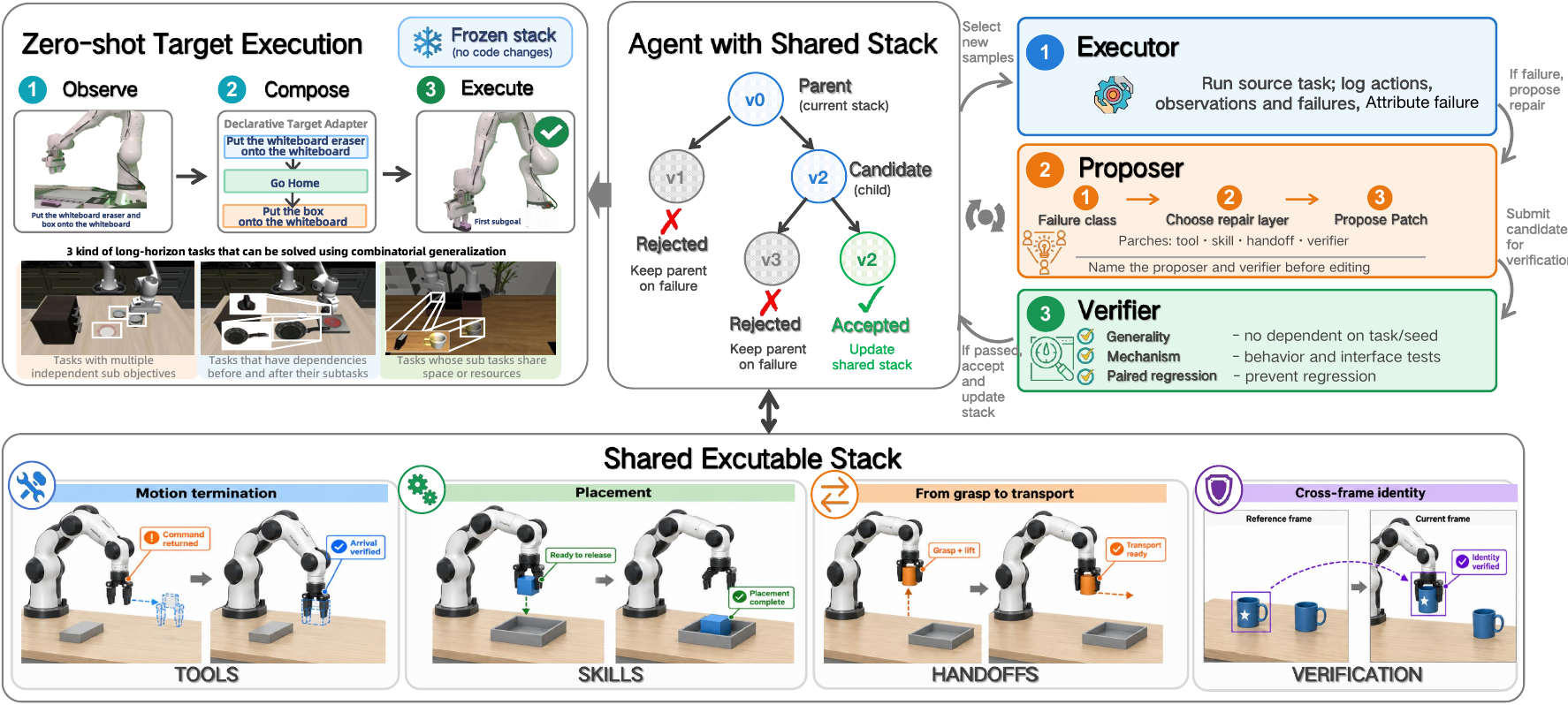}
\caption{Framework overview. The improvement loop operates on four executable levels. Both newly generated tools and repairs to existing components must pass the same acceptance process before entering the shared stack. Target execution does not modify the stack.}
\label{fig:overview}
\end{figure}
\section{Method}
\label{sec:method}

\subsection{Problem Formulation}
\label{sec:formulation}
Let $\mathcal{D}_{\mathrm{src}}$ denote the source dataset available for improvement and $\mathcal{D}_{\mathrm{tgt}}$ the target dataset used for frozen transfer. $\tau$ denotes one task. The goal is to obtain an executable shared stack through debugging with $\mathcal{D}_{\mathrm{src}}$.

After $k$ accepted modifications obtained from $\mathcal{D}_{\mathrm{src}}$, the executable stack is
\begin{equation}
\stack_k=(\mathcal{T}_k,\mathcal{K}_k,\mathcal{H}_k,\mathcal{V}_k),
\end{equation}
where $\mathcal{T}_k$ contains tools, $\mathcal{K}_k$ reusable skills, $\mathcal{H}_k$ handoff mechanisms, and $\mathcal{V}_k$ verifier mechanisms.
Learning and debugging occurs through changes and modifications to executable code; the foundation model is held fixed within a comparison. The declarative adapter $a_\tau$ provides object descriptions, support or destination descriptions, relation types, and goal dependencies.
A shared executor composes calls to these components using $a_\tau$ and the available observation history $h_t=(o_0,u_0,\ldots,o_t)$.

We use \emph{zero-shot compositional transfer} to mean that target-task compositions do not provide feedback for stack modification, and the frozen stack is executed without target-specific code repair. Source and target tasks may share objects, relations, or individual skills.

\subsection{Four-layer Shared Executable Stack}
\label{sec:stack}
\begin{figure}[!ht]
\centering
\includegraphics[width=0.6\linewidth,height=0.74\textheight,keepaspectratio]{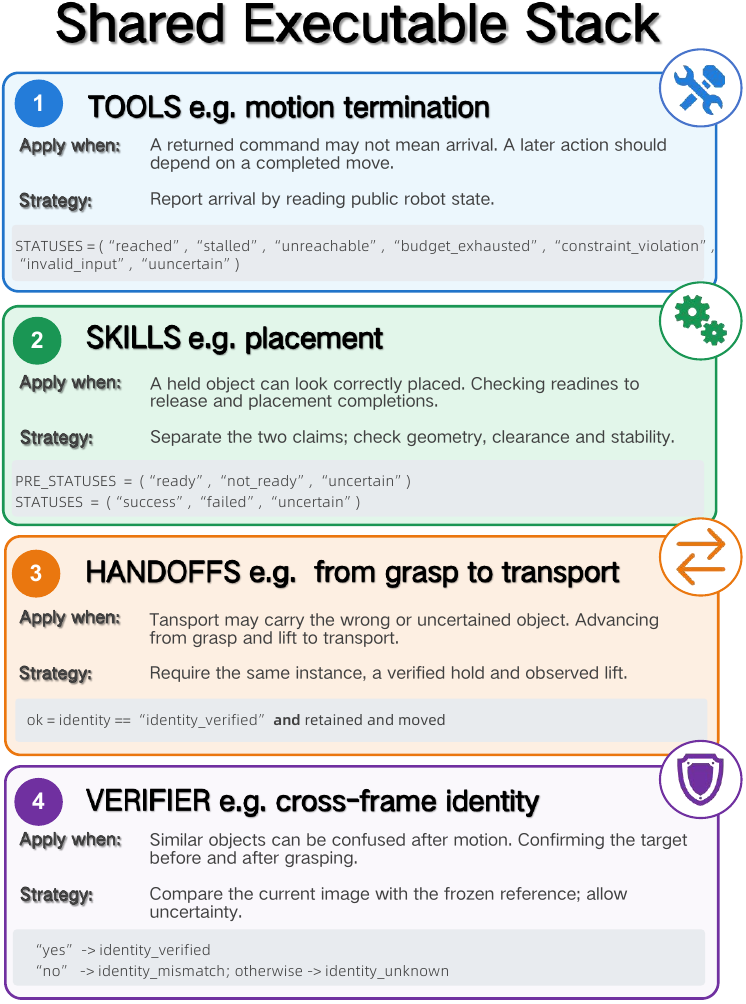}
\caption{Representative shared-stack interfaces.}
\label{fig:shared-stack-details}
\end{figure}

\paragraph{Tool.}
Tools expose reusable perception, geometry, planning, or execution operations. Their interfaces specify inputs, coordinate conventions, observable completion conditions, and resource limits.
The tool set is extensible. When an attributed source failure reveals that no available operation can satisfy a required capability, the agent proposes a new shared tool rather than embedding a task-specific workaround in the adapter. A proposal defines the missing capability, an interface, the permitted observations it consumes, and tests of its intended behavior. New tools are assembled from the permitted lower-level operations and are registered only after validation. Tool generation cannot introduce privileged sensing at deployment or alter the external evaluator. We distinguish these additions from repairs to existing tools in the experimental analysis. Figure~\ref{fig:shared-stack-details} illustrates representative interfaces.

\paragraph{Skill.}
A skill with execution contracts is represented by
\begin{equation}
K_i=(\operatorname{Pre}_i,\pi_i,\operatorname{Post}_i,B_i),
\end{equation}
where $\operatorname{Pre}_i$ defines observable prerequisites, $\pi_i$ is a parameterized program, $\operatorname{Post}_i$ specifies evidence required at completion, and $B_i$ bounds resource use. Preconditions and postconditions refer to task arguments and measured geometry, not identities of benchmark tasks. Execution returns evidence and a reason for termination, allowing the caller to distinguish a completed motion from an established task relation. Contracts are executable checks, not formal guarantees of correctness in unobserved states.

\paragraph{Handoff.}
The completion of one skill does not imply readiness for the next:
\begin{equation}
\operatorname{Post}_i(h_t)\not\Rightarrow\operatorname{Pre}_{i+1}(h_t).
\label{eq:handoff}
\end{equation}
A handoff updates the state used by the successor, checks its prerequisites, and checks previously achieved relations that later actions could disturb. Depending on the observed failure, the shared executor may re-observe, invoke a bounded recovery, or stop. Recovery must establish the missing prerequisite before execution resumes. For instance, a completed release motion requires a subsequent retreat before the next grasp becomes executable. The handoff makes this transition explicit.

\paragraph{Verifier.}
The verification layer evaluates object identity and task-relevant relations from permitted observations and execution history. The external evaluator independently measures its errors; the internal evaluator's own output is never used as the ground-truth success label. The external evaluator's role is distinct from a motion tool's arrival check: reaching a pose does not establish that the intended object is supported by the intended surface. The external verification uses support geometry and object identity across observations to decide whether the evidence permits advancement or recovery. 

Together, these levels define the unit of reuse. A validated tool can serve several skills, a skill can appear in several goal sequences, and a handoff or relation check can apply across tasks.

Figure~\ref{fig:shared-stack-details} shows four representative shared interfaces. The layers group responsibilities within one executable stack.

\subsection{Failure-Attributed Improvement Procedure}
\label{sec:repair}
The executor run source rollouts and records actions, observations and robot state. Source diagnosis use simulator ground truth to distinguish failure mechanisms, while the deployed stack remains restricted to permitted observations. The attribution procedure identifies the earliest supported violation of an execution requirement and records its stage, evidence, and affected component. It distinguishes an absent capability from an incorrectly implemented one.

The coding agent receives this evidence and proposes a candidate modification $\delta$. The proposal names the failure class, the stack level to modify, the intended repair, and the tests that would evaluate the repair. Examples in Figure~\ref{fig:shared-stack-details} include adding a missing constrained-motion operation, preventing unexecuted planning candidates from changing later solves, or revising a handoff that starts its successor before transport has completed.
The candidate repair modify an existing component of $\stack_k=(\mathcal{T}_k,\mathcal{K}_k,\mathcal{H}_k,\mathcal{V}_k)$ or extend the inventory. For instance, the candidate repair can extend the tool inventory as follows:
\begin{equation}
\stack' = \operatorname{Apply}(\stack_k,\delta),\qquad
\mathcal{T}'=\mathcal{T}_k\cup\{t_{\mathrm{new}}\}
\quad\text{for an accepted tool addition}.
\end{equation}
At each improvement iteration, the agent selects one shared-module slot for modification. Candidate changes are confined to that slot. For example, when the shared-tool slot is selected, the other shared components, diagnostic procedures, retrieval, search code, and data splits remain fixed during that iteration. First-failure attribution guides where to intervene, while the mechanism tests and paired execution determine whether that intervention works.

\recursia{} also allows the strategy update to change across iterations. Specifically, adjustments to the verification gate and the types of data used for debugging alter how candidate modifications are evaluated and which failures guide subsequent repair. These changes persist into later iterations, making the improvement procedure itself an object of adaptation. In later iterations, expanding patch selection from top-1 to top-$n$ effectively transforms the self-improvement process from a sequential chain into a beam search.

\subsection{Multidimensional Validation Gate}
\label{sec:validation}
Candidate acceptance combines three types of evidence. Firstly, generality checks reject benchmark-specific specialization in shared code, including task or seed identifiers and object- or scene-specific control branches. Geometric constants require a documented source, such as robot calibration, measured geometry, or a declared object-independent construction. 

Secondly, mechanism tests check the proposed behavior before broader execution. Perception and verification patches can be replayed on recorded observations. Motion patches require execution tests, because an old trajectory does not reveal what a new controller would do. Tests also check interface compatibility and whether downstream actions respect failure returns. A newly generated tool must demonstrate both its intended capability and bounded handling of unsupported requests.

Finally, parent and candidate are compared on identical source-task states under a resource policy fixed before candidate evaluation in the current iteration. Let $G(\delta)$ indicate the generality checks, $U(\delta)$ the mechanism tests, and $R(\stack',\stack_k;\mathcal{D}_{\mathrm{sel}},B)$ the paired performance and regression gate. Acceptance is
\begin{equation}
\operatorname{Accept}(\delta)=G(\delta)\land U(\delta)
\land R(\stack',\stack_k;\mathcal{D}_{\mathrm{sel}},B).
\label{eq:accept}
\end{equation}
The gate considers source-task success, regressions on other source tasks, and resource use. At the beginning of each RSI iteration, the coordinator fixes the evaluation panels, acceptance criteria, and resource budgets for that iteration. These settings may change between iterations as the stack evolves, but remain fixed when comparing the parent with candidates within the same iteration. Rejected candidates do not replace the parent. Accepted candidates retain their code hash, source failure, tests, and paired results. 

\begin{algorithm}[t]
\caption{\recursia{} Debugging Phrase for One Iteration}
\label{alg:rsi}
\begin{algorithmic}[1]
\Require Initial stack $\stack_0$; source development task and selection tasks; budget $B$
\State $k\gets 0$
\For{each source-task improvement step within budget}
    \State Execute $\stack_k$ with a source-task adapter
    \State Attribute the first supported failure from aligned source traces
    \State Identify a missing capability or a faulty shared component
    \State Propose $\delta$: generate a tool or repair the selected stack level
    \State $\stack'\gets\operatorname{Apply}(\stack_k,\delta)$ in an isolated candidate version
    \If{generality checks and mechanism tests pass}
        \State Run parent-candidate paired development and selection tests
        \If{Multidimensional validation gates pass}
            \State $\stack_{k+1}\gets\stack'$; archive repair evidence; $k\gets k+1$
        \Else
            \State Retain $\stack_k$ and archive the rejection
        \EndIf
    \EndIf
\EndFor
\State Inference: Execute target adapters using the frozen stack without target-driven repair
\end{algorithmic}
\end{algorithm}

At inference time, all shared implementations are frozen. The target adapter specifies the requested composition, and execution logs identify the versions invoked by each subgoal. 

\section{Experiments}
\label{sec:experiments}
Our experiments examine whether source-side improvement of a shared executable stack transfers to unseen long-horizon compositions and which levels of the shared stack contribute to transfer.

\subsection{Experimental Setup}
\label{sec:setup}

\paragraph{Setup.}
We use LIBERO-90~\citep{libero2024} for debugging, including failure diagnosis, repair generation, and candidate selection. Improvements persist in a versioned executable stack rather than in target-specific programs. All source-side development is completed before target evaluation, and the selected stack is frozen together with its tools, skill implementations, handoff logic, and verifiers. The coding model and its parameters remain fixed within each controlled comparison. 

The first track evaluates the frozen stack on
LIBERO-Pro Long, following the long-horizon transfer setting studied by ASPIRE~\citep{zhou2025liberopro,lu2026aspire}, with 10 long-horizon tasks evaluated along two axes. \emph{Pos} changes initial object positions; \emph{Task} changes the task specification. Each task--axis pair is evaluated on seeds 1--50, yielding 500 episodes per axis. The 10 task names and their axis-specific results are listed in Appendix~\ref{app:zero-shot-per-task}. 
The second track evaluates on a physical Franka FR3 equipped with a Franka Hand. Physical evaluation trajectories are not used for stack repair or selection.

\paragraph{Baselines.}
We compare with OpenVLA~\citep{kim2024openvlaopensourcevisionlanguageactionmodel}, $\pi_0$~\citep{black2024pi0visionlanguageactionflowmodel}, $\pi_{0.5}$~\citep{intelligence2025pi05}, CaP-Agent0~\citep{capx2026}, and ASPIRE~\citep{lu2026aspire}. CaP-Agent0's published protocol regenerates programs per seed with test-time reasoning and retries, whereas ASPIRE fixes one program per task without target debugging; this difference is retained when interpreting the comparison. 

\paragraph{Metrics.}
We measure full-task success using the official benchmark evaluator. For task $i$ and perturbation axis $a$, success is the fraction of successful episodes among its 50 evaluation states. Axis success is the unweighted mean over the 10 task rates; Overall is the mean of the \emph{Pos} and \emph{Task} macro-averages. Rates are reported on $[0,1]$. Failures, timeouts, and program errors remain in the denominator. We report source-improvement costs separately from target-execution costs, including model input/output tokens, wall time, and perception calls. Matched reruns use identical initial states and report paired outcomes. 

\subsection{Evaluation in Simulation: Zero-Shot Transfer to Long-Horizon Tasks on  LIBERO-Pro-Long}
\label{sec:transfer}
\begin{figure}[t]
\centering
\includegraphics[width=\linewidth]{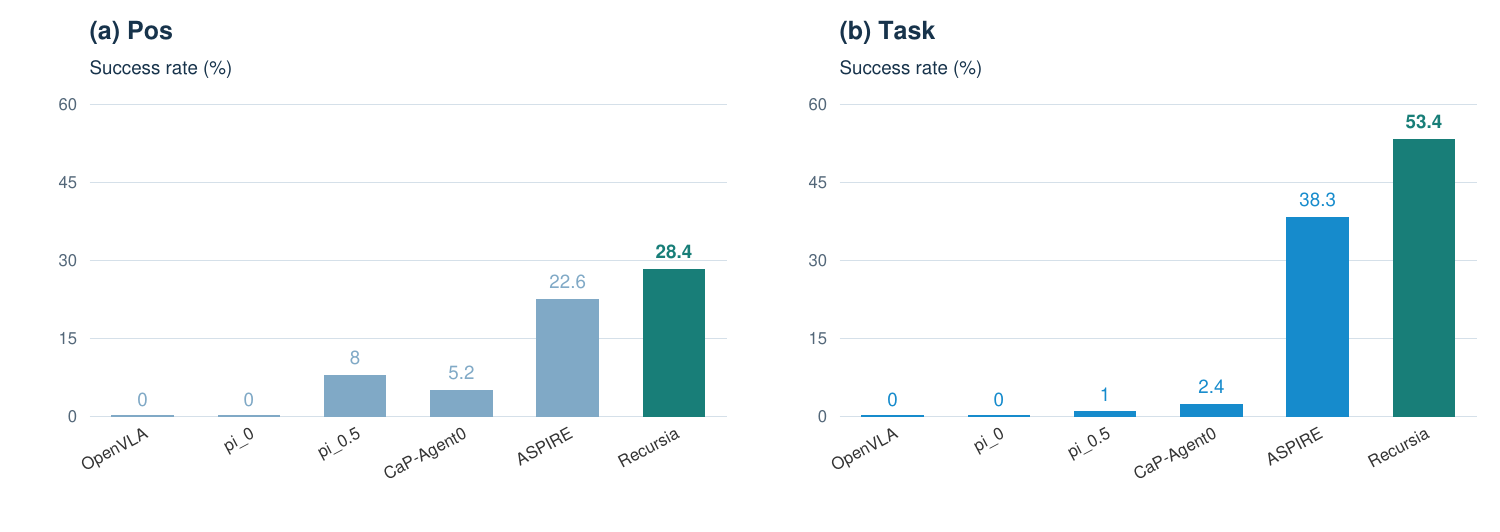}
\caption{Zero-shot transfer to LIBERO-Pro Long. (a) Position perturbations (\emph{Pos}). (b) Task-specification perturbations (\emph{Task}). Baseline success rates are reproduced from the existing comparison; \recursia{} achieves success rates of 0.284 on \emph{Pos} and 0.534 on \emph{Task}. Aggregate and per-task tables appear in Appendix~\ref{app:zero-shot-per-task}.}
\label{fig:transfer}
\end{figure}

Here $N$ denotes the number of improvement iterations; each iteration processes five source samples.
Figure~\ref{fig:transfer} presents the primary zero-shot comparison separately for position (\emph{Pos}) and task-specification (\emph{Task}) perturbations. The source-experience scaling analysis is provided in Appendix~\ref{app:transfer-scaling}, Figure~\ref{fig:transfer-scaling}. We report separately frozen \recursia{} stacks after $N\in\{0,3,6,9,10,12,13,14,15,16,17,18\}$ improvement iterations to examine how transfer changes with source-side experience. Each snapshot is selected using source data only and evaluated without target-driven modification. This comparison tests whether source-side changes to shared execution mechanisms remain useful when the target composition changes and no target feedback is available for repair.  \recursia{} achieves success rates of 0.284 on \emph{Pos} and 0.534 on \emph{Task}. Across the supplied scaling checkpoints, success rises from 0 at $N=0$ to 0.232 and 0.434 at $N=18$, respectively, with nonmonotonic intermediate changes.

\subsection{Evaluation on a Physical Robot: Zero-Shot Transfer to Long-Horizon Tasks with Franka FR3}
\label{sec:realrobot}

We compare \recursia{} with Skill Document Only (SD-only)
on a Franka FR3 equipped with a Franka Hand.
SD-only uses the tool implementation of ASPIRE and  updates a single natural-language skill document,
whereas \recursia{} updates a shared executable Python stack.
Both use GPT-6 Astra, front and wrist observations, and
the same six robot tools with SAM3 and Contact-GraspNet.

The six physical source tasks cover placing a mug or milk
carton in a basket, placing white and blue mugs on the left
and right plates, placing a plastic bowl on a plate, and
placing a mug in the bowl.
Each source task contributes three retained development
rollouts. SD-only runs three passes through the source suite;
\recursia{} groups the tasks into three two-task chunks,
evaluating the current stack and two candidate updates
on each pair.

The targets compose these actions:
T1 places the mug and milk carton in the same basket;
T2 places the blue mug on the left plate and the white mug
on the right plate, reversing their source assignments;
T3 first places the bowl on the plate, then the mug in
the relocated bowl.
Source knowledge is frozen before evaluation.
Within-trial feedback corrections do not update persistent
knowledge. Each target is evaluated ten times per method,
with at most 20 code blocks per trial.
Task success is assessed manually.

\begin{table}[t]
\centering
\small
\caption{Success rates (\%) on physical compositional
target tasks using Franka FR3 with Franka Hand.}
\label{tab:realrobot}
\setlength{\tabcolsep}{12pt}
\begin{tabular}{lrr}
\toprule
Task & SD-only & \recursia{} \\
\midrule
T1: Shared basket        & 0.0  & 20.0  \\
T2: Reassigned plates    & 60.0 & 80.0  \\
T3: Bowl--mug dependency & 20.0  & 100.0 \\
\midrule
Average                  & 26.7 & 66.7  \\
\bottomrule
\end{tabular}
\end{table}

\subsection{Analysis of Compositional Generalization}
\label{sec:composition}
The \recursia{} framework targets compositional generalization, combining knowledge from multiple source tasks to enable zero-shot transfer to long-horizon target tasks. This subsection aims to demonstrate this capability through retrospective case studies. Compositional transfer requires both competent constituent skills and compatible states at their boundaries. 
To disentangle improvements in individual skills from those in compositional ability, we examine this distinction through retrospective execution cases, separating two events: completion of the predecessor's goal and establishment of the successor's prerequisites. 
In particular, while the predecessor task is complete in the current state, this does not guarantee that the next object can be readily identified, reached, or manipulated.
A useful diagnostic is the difference between isolated and chained success:
\begin{equation}
\Delta_{\mathrm{chain}}=p_{\mathrm{success}}^{\mathrm{isolated}}
-p_{\mathrm{success}}^{\mathrm{chained}}
\end{equation}
This quantity measures sensitivity induced by composition capability. 

\paragraph{Improving isolated skills does not remove the downstream bottleneck.}
At $N=11$, the standalone stove-on operation improved from $0/10$ to $10/10$ on both development and selection datasets, whereas the stove-pan composition remained $0/10$. In the inspected development rollouts, the stove-on predicate became \textit{True}, but subsequent manipulation lost the pan after a grasp attempt; alternative candidates were also rejected as outside the inverse-kinematics workspace. Thus, the rotary repair improved the first operation without establishing the conditions needed to complete the second. This comparison locates the remaining bottleneck and shows the difference between isolated and chained success. Our approach filled this gap by handoff repair mechanism, in the later iterations. 

\begin{figure}[t]
\centering
\includegraphics[width=0.8\linewidth]{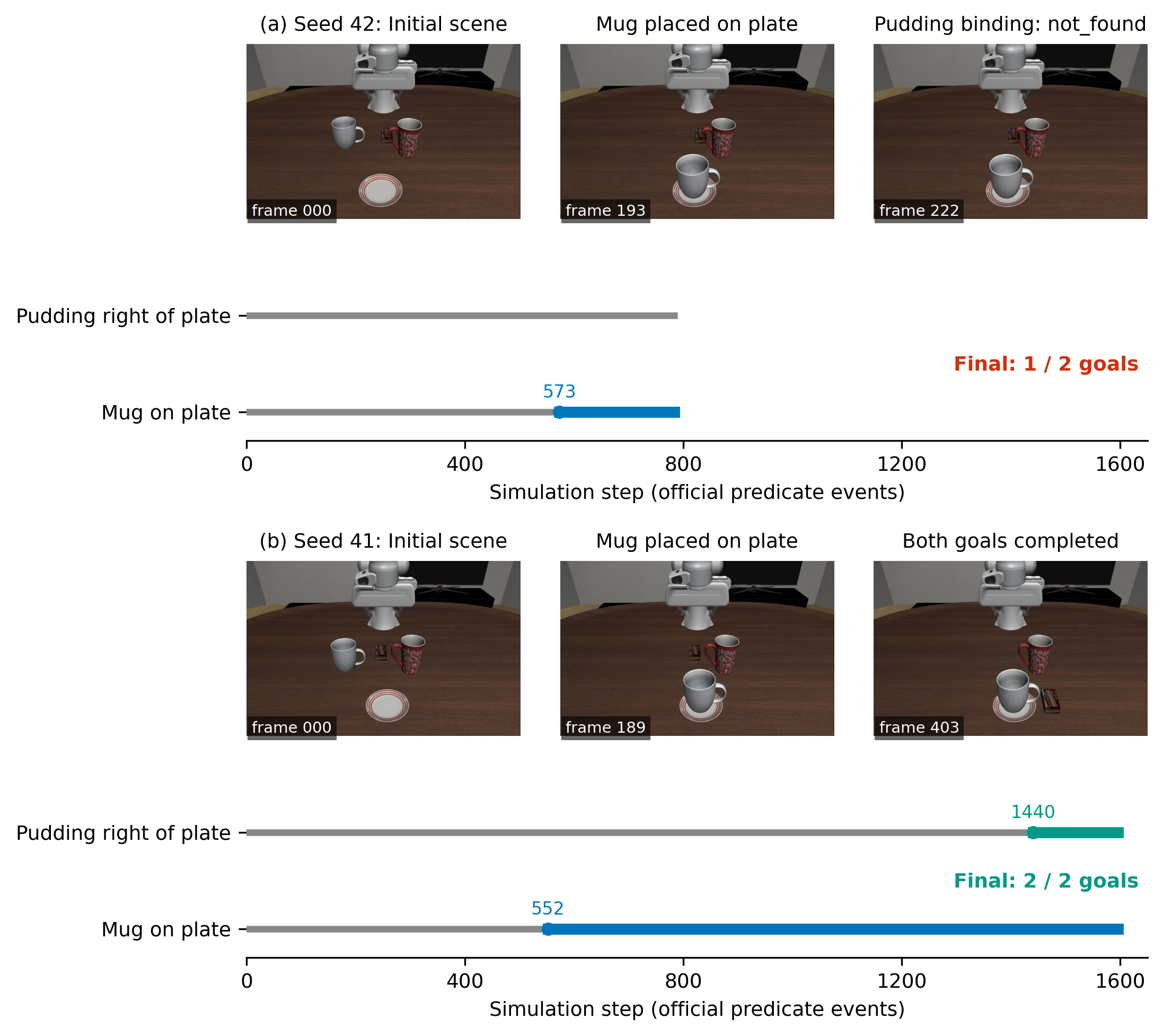}
\caption{
Recorded compositional generalization cases. Original RGB recordings and official predicate timelines for the N=6 mug--pudding sequence. (a) Seed 42 completes the mug goal, but pudding binding returns \texttt{not\_found}; the earlier goal remains true through termination. (b) Seed 41 completes both goals while preserving the first. Gray segments denote false predicates; blue and teal segments denote true predicates. Across ten swap-suite episodes, the mug predicate is true at termination in all ten, whereas the full conjunction succeeds in one. The two illustrated seeds have different initial states. RGB frame identifiers and simulation-step timestamps use separate counters. 
}
\label{fig:handoff}
\end{figure}

\paragraph{Relational success and successor readiness are distinct.}
In the $N=6$ mug-pudding sequence, the mug-on-plate predicate became \textit{True} in all ten \emph{Pos} episodes, while the full conjunction succeeded in only one. In seed 42, the mug goal became \textit{True} at step 573 and remained \textit{True}, but subsequent pudding binding returned \texttt{not\_found}. Seed 41 provides a successful contrast: the mug goal became \textit{True} at step 552 and the pudding goal at step 1440, with the first goal preserved (Figure~\ref{fig:handoff}). A drawer sequence exhibits the same failure structure: seed 48 achieved bowl containment, but closing stopped at \texttt{part\_not\_found}. These traces identify perceptual binding as a prerequisite that is not certified by predecessor goal completion.

Together, these cases distinguish failures of successor execution from failures to establish successor readiness. The observed mug-pudding failure belongs to the second category, with the first goal preserved; the stove-pan failure exposes downstream manipulation limitations after stove-on succeeds. These evidence supports a conclusion: predecessor completion and successor readiness must be examined separately and processed by the handoff mechanism.

\subsection{Analysis of Inference Methods}
\label{sec:inference-strategies}
\begin{figure}[t]
\centering
\includegraphics[width=\linewidth]{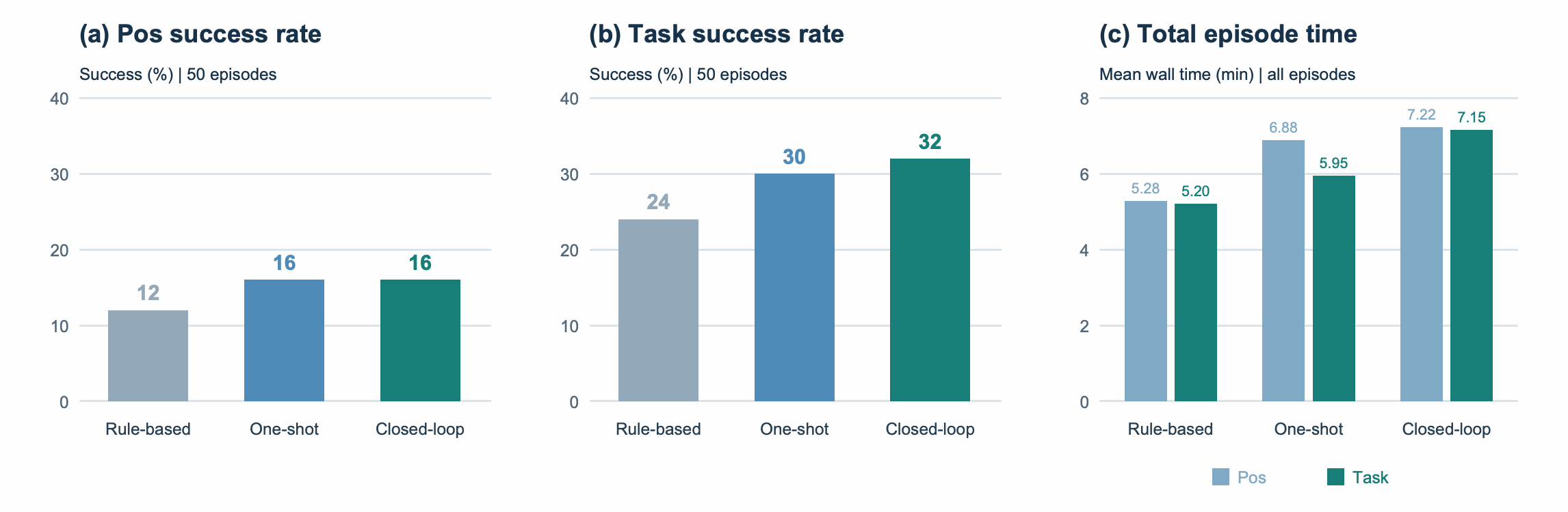}
\caption{Inference-driver comparison on a frozen stack of the $N=13$ iteration. (a) \emph{Pos} and (b) \emph{Task} success rates, each 50 matched episodes. (c) Mean end-to-end episode time, including program generation and robot execution, for all episodes.}
\label{fig:inference-strategies}
\end{figure}

All inference methods are driven by declarative adapters. 
We compare three inference drivers over the same frozen shared stack: rule-based driver, one-shot task-program driver, and online closed-loop driver. These drivers differ in how high-level execution decisions are specified and updated. The rule-based driver follows predefined decision rules. The one-shot task-program driver executes an initially generated program, which may include observation-dependent branches and bounded retries. The online closed-loop\footnote{The label ``closed-loop'' here denotes the inference driver's decision process, rather than a property exclusive to its low-level robot controller.} driver conditions subsequent decisions on observations and execution outcomes throughout the episode. All three may use observation feedback and closed-loop low-level control. The results are shown in Figure~\ref{fig:inference-strategies}.
One-shot generation describes the lifetime of the program, rather than an absence of feedback during its execution. In ASPIRE's zero-shot setting, task-program generation excludes target-side debugging, but the one-shot generated program may still re-observe the scene and use closed-loop low-level control~\citep{lu2026aspire}. Here, we use the frozen shared stack.

\paragraph{The online closed-loop driver outperforms the one-shot task-program driver.}
The comparison between the online closed-loop driver and one-shot task-program driver is more limited. After multiple tries, the online closed-loop driver always gains more successful trajectories. The trajectories show that the drivers have different failure and success profiles. It reveals that the closed-loop decision making may resolve failures encountered by a fixed generated program while introducing other failures.

\paragraph{Feedback incurs latency without uniform gains across perturbation axes.}
The rule-based, one-shot, and closed-loop drivers achieve \emph{Pos} success rates of $12\%$, $16\%$, and $16\%$, and \emph{Task} success rates of $24\%$, $30\%$, and $32\%$, respectively. Mean end-to-end episode times are $5.28$, $6.88$, and $7.22$ minutes on \emph{Pos}, and $5.20$, $5.95$, and $7.15$ minutes on \emph{Task} (Figure~\ref{fig:inference-strategies}). These durations include generation and execution, with runtime model latency counted within execution; failures and timeouts remain in the averages.

\paragraph{Feedback-condition mechanism incurs latency without uniform gains across perturbation axes.}
Mean end-to-end episode times are $5.28$, $6.88$, and $7.22$ minutes on \emph{Pos}, and $5.20$, $5.95$, and $7.15$ minutes on \emph{Task}. These durations include generation and execution, with runtime model latency counted within execution; failures and timeouts remain in the averages.
Relative to one-shot generation, closed-loop inference adds approximately $21$ seconds per \emph{Pos} episode, and $72$ seconds per \emph{Task} episode for a two-percentage-point gain. The incremental benefit is therefore concentrated in \emph{Task} perturbations, whereas the \emph{Pos} results do not show an aggregate improvement from additional feedback-conditioned decisions. Relative to the rule-based driver, one-shot generation gains four and six percentage points at additional mean costs of approximately $96$ and $45$ seconds on \emph{Pos} and \emph{Task}. In this comparison, much of the observed success improvement is already obtained by one-shot orchestration.

\subsection{Analysis of the Validation Gate}
\label{sec:gates}
The \recursia{} framework incorporates two critical validation components for the validation gate, i.e., the \textit{selection gate} and the \textit{regression gate}, to address two distinct risks: a candidate may correct a local mechanism without improving overall task completion (mitigated by the selection gate), or improve aggregate completion while disrupting previously successful states (mitigated by the regression gate). This subsection uses retrospective case studies to demonstrate their effectiveness. In the book-transfer task, the development summary reported candidate-filter rejection in eight of ten episodes for early iterations $N=8$. Inspecting the earlier execution revealed a defect: the robot had already grasped and lifted the intended book, but a post-lift visual query selected a similar book that remained on the table. The low re-detected position was treated as evidence of a failed lift, causing the gripper to open and subsequent grasp attempts to fail. 
The subsequent iterations attempt to repair this identity check. We employ the validation framework to ensure the recognition of a known stationary competitor following a verified lift, while preserving the existing motion and hold checks.

\paragraph{Selection gate rejects candidates with insufficient improvement on the selection dataset.}
A repair during $N=9$ changed the transport-to-descent boundary for book insertion so that entry into the opening remained under contact-aware control. Development release refusals attributed to the lateral corridor decreased from 8 to 2, and most tests passed. Nevertheless, official completion improved only from $0/10$ to $1/10$ in the development dataset and remained $0/10$ in the selection dataset, below the predeclared gains of two and one successes, respectively. The reason is that removing a prominent failure mechanism exposed remaining arrival, contact, and grasp failures. Although the candidate passed the paired-loss checks, it was rejected by the selection gate for insufficient improvement on the selection dataset.

\begin{figure}[t]
\centering
\includegraphics[width=0.8\linewidth]{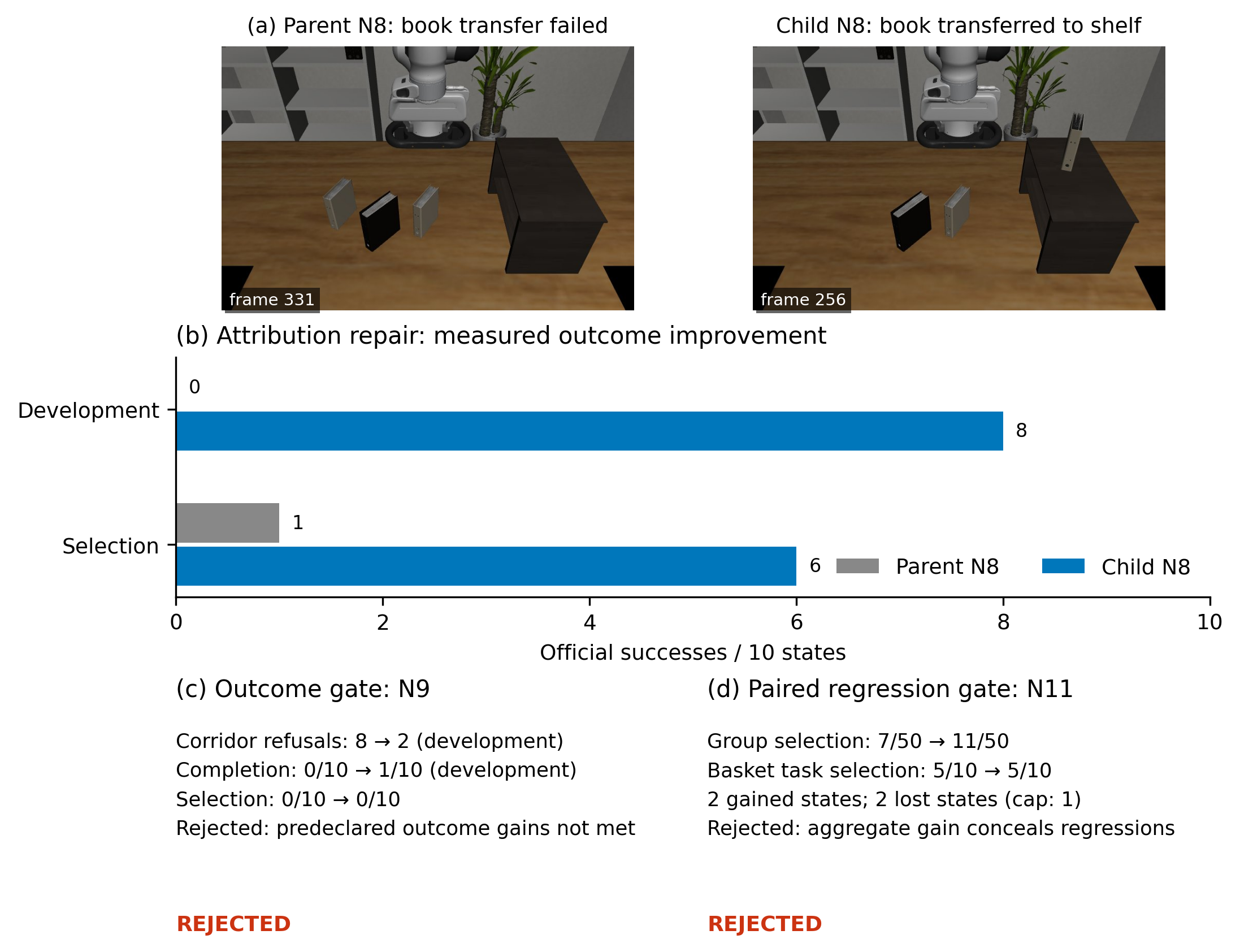}
\caption{Recorded validation-gate cases. (a) Initial state. Original simulator frames from the N=8 iteration, both with seed 86. (b)  Development and selection successes for this task. (c) The iteration N=9 reduced a local failure mechanism but failed the predeclared outcome gains. (d) The iteration N=11 improved the aggregate selection score while losing two previously successful basket-placement states, exceeding the per-task loss cap.}
\label{fig:validation_gates}
\end{figure}

\paragraph{Regression gate prevents aggregate gains from concealing lost successes.}
The N=11 iteration increased development completion from $2/50$ to $3/50$ and selection completion from $7/50$ to $11/50$. However, one basket-placement task retained its selection total of $5/10$ by gaining two states and losing two previously successful states. The two losses exceeded the per-task cap of one, and the candidate was rejected by the regression gate. Development book-insertion corridor refusals also fell from nine to zero without producing a completed insertion. These cases motivate checking trajectories, outcomes and paired losses in addition to aggregate development scores (Figure~\ref{fig:validation_gates}).

\subsection{Analysis of Tool Repair and Tool Generation}
\label{sec:provenance}
Table~\ref{tab:repairs} summarizes representative entries and shows generated repairs extend the callable inventory.
The N=10 iteration added support-instance and placement semantics for previously refused \texttt{front/middle} relations, reusing the existing instance selector and placement primitives. Instruction dispatch coverage increased from $81/90$ to $84/90$, and the \texttt{front-of-mug} task improved from $0/10$ to $8/10$ in development dataset and $0/10$ to $7/10$ in selection dataset. The N=11 iteration subsequently generated a shared \texttt{run\_rotary} driver that served both turn-on and turn-off instructions, corrected the pinch frame, and re-anchored turn increments on the executed tool pose. Twenty tests passed, and the driver was invoked in both standalone stove activation and the stove--pan sequence.

\begin{table}[t]
\centering
\small
\setlength{\tabcolsep}{4pt}
\begin{tabular}{@{}p{0.10\linewidth}p{0.26\linewidth}p{0.29\linewidth}p{0.28\linewidth}@{}}
\toprule
\textbf{Entry} & \textbf{Failure} & \textbf{Repair} & \textbf{Recorded Result} \\
\midrule
N=10 & Lack of front/middle reference relations & Shared instance selection and relation dispatch & Dispatch $81/90\!\rightarrow\!84/90$; gains measured on source tasks. \\
N=11 & Rotary localization, pinch, and arc defects & Shared turn-on/off driver & Turn-on $10/10$ in each panel. \\
\bottomrule
\end{tabular}
\caption{Representative repairs. Rates refer to the specified tasks, not the complete Long-Pro benchmark.}
\label{tab:repairs}
\end{table}

\section{Related Work}
\label{sec:related}

\paragraph{Embodied Agents.}
Language-guided robot agents connect high-level reasoning to physical execution through learned action models or executable programs. Vision-language-action models (VLA) such as RT-2, OpenVLA, and $\pi$ series map visual observations and language instructions to robot actions~\citep{brohan2023rt2,kim2024openvlaopensourcevisionlanguageactionmodel,black2024pi0visionlanguageactionflowmodel}. Programmatic approaches instead express control through calls to perception and action interfaces. Code as Policies composes these calls into robot programs and recursively generates missing helper functions~\citep{liang2023code}, while ProgPrompt grounds plan generation in programmatic descriptions of available actions and objects~\citep{singh2023progprompt}. CaP-X extends this setting with systematic evaluation of coding agents, structured execution feedback, and automatic skill synthesis~\citep{capx2026}. Together, these approaches provide the building blocks for language-guided manipulation. \recursia{} focuses on how those building blocks improve across tasks: it persistently modifies the shared tools, skills, handoffs, and verifiers that generated programs invoke.

\paragraph{Embodied Recursive Self-Improvement.}
A central question in recursive self-improving agents is where experienc and knowledge is retained and how it changes subsequent behavior. The Darwin G\"odel Machine retains improvements in the agent's own code and evaluates successive variants on coding benchmarks~\citep{zhang2025dgm}. In robotics, Eureka uses language models to generate reward functions for policy learning~\citep{eureka2023}, while evolutionary code search directly synthesizes executable control policies~\citep{guo2026codeevolution}. Skill-library approaches retain experience as reusable procedures. Voyager accumulates executable skills in Minecraft~\citep{voyager2023}; Lifelong Robot Library Learning bootstraps composable skills for robot control~\citep{tziafas2024lrll}; and Uni-Skill requests new implementations when its existing repository cannot support a plan~\citep{uniskill2026}. These approaches make accumulated experience available to later tasks through improved policies, programs, or skill repertoires.
ASPIRE~\citep{lu2026aspire} brings execution-driven repair and skill accumulation into a continual robot-programming loop. It uses execution traces to diagnose failures, distills validated fixes into a transferable skill library. Its evaluation includes transfer to unseen long-horizon tasks, making it the closest precursor to \recursia{}. Zero-shot transfer is also studied in the context of lifelong learning~\citep{zhou2025lifelong}, where a robot learns to perform new tasks without additional training. 
\recursia{} differs from these approaches by focusing on the shared executable stack as the unit of improvement, allowing for more systematic and reusable repairs across tasks.

\paragraph{Compositional Generalization for Robotics.}
Compositional generalization requires familiar capabilities to remain effective when objects, goals, or action sequences are combined in new ways. CLIPort studies language-conditioned manipulation through semantic and spatial abstractions~\citep{shridhar2022cliport}, and VIMA evaluates progressively more novel combinations of task components~\citep{jiang2023vima}. At longer horizons, CALVIN evaluates sequences of language-conditioned manipulation tasks~\citep{mees2022calvin}, while LIBERO studies knowledge transfer across spatial, object, goal, and task variations~\citep{libero2024}. LIBERO-PRO adds perturbations to initial object positions and task specifications~\citep{zhou2025liberopro}. These evaluation settings expose several dimensions of transfer, from recognizing a new combination of familiar concepts to executing a sequence under changed conditions.

Reliable sequence execution also depends on the states passed between skills. BATON addresses this dependency by exploring subtasks separately and storing transition-aware memory: verification checks readiness, handoff restoration repairs entry conditions, and lookahead favors outcomes that the next subtask can use~\citep{xu2026baton}. Its treatment of transitions makes explicit why a skill can succeed in isolation yet fail as part of a chain. \recursia{} places this dependency within a broader stack-improvement problem. Handoff repairs are validated alongside changes to tools, skill contracts, and verification, and their shared implementations are reused unchanged on target compositions. Our evaluation therefore examines both the competence of individual skills and the reliability of their composition, tracing transfer to source-side changes in the executable stack.

\section{Limitations and Conclusion}
\label{sec:conclusion}
Our evaluation has several limitations. First, source-side ground-truth diagnosis provides information that may not be available on a physical robot. In simulation, failed episodes can be reset to a controlled initial state and replayed repeatedly. On a physical robot, resetting the scene can be slow, unsafe, or impossible after an object has moved, and the diagnostic evidence needed to identify the first violated requirement is often incomplete or expensive to obtain. Deployment does not consume simulator ground truth, but transferring the improvement procedure itself to real hardware therefore requires reliable alternatives, such as calibrated state logging, external sensing, human annotations, or conservative uncertainty handling. 
Second, we have not evaluated cross-embodiment generalization. All source and target experiments use the same robot embodiment and execution interfaces. A shared repair may depend on kinematics, gripper geometry, controller conventions, or calibration details, so transfer across robot platforms remains untested. 

We present a framework for improving the shared executable mechanisms through which coding agents compose robot skills. The stack can grow when a required tool is missing and change when existing tools, contracts, handoffs, or verification fail. Source failure attribution directs these changes, while validation gates determine which changes persist. Frozen reuse then connects a source-side modification to the code executed in an unseen long-horizon composition. The primary LIBERO-Pro Long results are 28.4\% on \emph{Pos} and 53.4\% on \emph{Task}. 
\bibliography{references}
\bibliographystyle{plainnat}
\clearpage
\appendix

\section{Debugging Pipeline}
\label{app:debugging-pipeline}
This appendix documents the partial artifact, including the implementation description. 
The debugging pipeline separates proposal generation from evaluation. The verifier runs a frozen parent, records public observations and simulator/physical  diagnostics, and supplies development evidence to one proposer. The proposer diagnoses a failure, makes several proposals, modifies shared code, replays the proposed decisions offline, and supplies tests. The verifier builds only after the final report, checks the candidate, runs a fixed paired plan, and records acceptance or rejection. An accepted program becomes the next parent; a rejected candidate leaves the parent unchanged.
The prompt boxes reproduce templates and partial excerpts. Deployment-specific paths and login fields are replaced by \texttt{\$ASPIRE\_ROOT}, \texttt{<user>}, and \texttt{<server>}; typographic punctuation is normalized. Braced variables are filled by the coordinator. These are implementation records, including their historical restrictions.

\subsection{Shared-Stack Proposer}
The proposer receives the frozen program, running log, failure attribution, descriptor data, and patch-building interface. Its output is one shared-module change, with diagnosis and offline evidence preceding simulation. The nine builder module labels are \texttt{motion\_preconditions}, \texttt{grasp\_geometry}, \texttt{hold\_contract}, \texttt{place\_primitive}, \texttt{verifier}, \texttt{binding}, \texttt{language\_adapter}, \texttt{manipulation\_primitive}, and \texttt{task\_driver}. These labels support attribution and dependency checks across the four conceptual layers.

\begin{promptbox}{Shared-Stack Proposer}
\textbf{Role.} You are the proposer in a recursive self-improvement loop. Repair one reusable component of a shared robot program from source-task failures.\par

\textbf{Evidence.} You receive the following message:
\begin{itemize}
    \item Frozen parent program
    \item Development failure attribution
    \item Recorded public observations
    \item Descriptor data
    \item Patch interface
\end{itemize}

Simulator truth may be used to diagnose failures and evaluate offline replay, but must never enter the executable policy.\par

\textbf{Procedure.} Choose one failure class that a coherent shared change can address across the development tasks. Identify the first violated requirement and the responsible module. Write the diagnosis before the patch. Replay the parent and candidate decisions on the recorded observations, and check that previously passing decisions remain unchanged. Derive new constants from measured geometry or documented robot quantities.\par

\textbf{Constraints.} Produce one task-agnostic patch to one declared module. Declare all affected dependencies. Use public robot and perception APIs. Read task facts from descriptors rather than embedding task, object, scene, color, or seed identifiers. Do not run policy episodes or tune constants to episode outcomes. Write only within the assigned proposal workspace.\par

\textbf{Deliverables.} Return the diagnosis, patch, offline replay and its results, and meaningful tests. The final report must state the failure class, modified module, proposed mechanism, evidence, test results, unresolved limitations, and artifact locations. The evaluator builds the candidate only after this report.\par

\end{promptbox}

\subsection{Verifier}
The ``verifier'' role is implemented by deterministic checks and an evaluator review. The coordinator runbook specifies how to invoke the builder, coverage gate, paired rollout gate, and review, and when the accepted parent may change. The adapted instructions below summarize the coordinator's evaluation responsibilities.

\begin{promptbox}{Candidate validation}
\textbf{Role.} You are the coordinator and evaluator. Control candidate evaluation, and acceptance while keeping the current parent frozen.\par

\textbf{Proposal.} Predeclare the parent, missing pairs, and development tasks. Provide only development evidence to one proposer. Resume an interrupted proposal rather than generating a duplicate. Wait for its diagnosis, replay, tests, and final report.\par

\textbf{Validation.} Build the final patch through the fixed builder and run the coverage check. A build or coverage failure consumes the proposal slot and stops candidate evaluation. Otherwise, freeze the candidate and evaluation inputs, then run the predeclared paired plan. Check complete episode accounting, policy, matched initial-state identities, and runtime integrity.\par

\textbf{Review.} Apply the capability and regression gates. Inspect new successes for genuine physical completion and inspect new regression failures. Numeric passage alone is insufficient. Do not disclose selection or hidden regression details to the proposer.\par

\textbf{Decision.} Record acceptance or rejection and its evidence. Update the parent only after all required checks and evaluator review pass. On rejection, retain the parent. At the declared stopping point, snapshot the last accepted program and record its lineage.\par

\end{promptbox}

\paragraph{Check implementation.}
The builder checks exact patch anchors, the declared module, a forbidden-literal list in inserted text, Python compilation, and supplied \texttt{test\_*.py} files. Missing tests produce a warning in the inherited builder; the protocol and evaluator require them.

\paragraph{Paired acceptance.}
Each rollout stage contains {\texttt{NUM}} rows: {\texttt{NUM}} development episodes (seeds 86--95), {\texttt{NUM}} selection episodes (the same tasks, seeds 76--85), {\texttt{NUM}} ordinary-source regression episodes, and {\texttt{NUM}} synthetic-regression episodes. Here ``synthetic'' denotes instantiated simulator tasks in the protocol. A group's net official gain must be at least two development states and one selection state. Losses are counted separately: at most one state per task per role, and at most one in each regression role. Both stages must pass integrity checks; at most four clean horizon timeouts are permitted. Passing the numeric gate is insufficient: the evaluator must inspect new physical completions and new regression failures.

\section{Inference Pipeline}
\label{app:inference-pipeline}
Inference executes a frozen shared program. The data-only adapter \texttt{adapter\_v3.py} constructs descriptors from the public runtime instruction \texttt{env.handle.task\_language}. To keep general, it does not take a task identifier, scene identifier, seed, or simulator state as input. The program's \texttt{la\_parse} produces ordered subgoals and \texttt{la\_dispatch} checks every subgoal before any motion. Unsupported instructions stop explicitly. These functions are deterministic code.
The following boxes present concise versions of its localization and identity prompts, using symbolic placeholders for runtime values.

\subsection{Object Localization}
Revision {\texttt{NUM}} preserves a confident text-based binding and uses VLM localization when text binding fails. The model sees a public camera image, the instruction, the current object phrase, and the execution phase. It returns a pixel or abstains. The point prompts SAM3; masks must contain the point, occupy no more than 8\% of the image, and yield at least 50 deprojected depth points. Only \texttt{medium} or \texttt{high} confidence proceeds.

\begin{promptbox}{Object localization | Initial reference}
\textbf{Context.} Task instruction: \texttt{\{instruction\}}. Object phrase: \texttt{\{phrase\}}. Execution phase: \texttt{\{phase\}}.\par

\textbf{Image.} Image 1 is the current camera view, \texttt{\{W\}} by \texttt{\{H\}} pixels. Coordinates start at the upper left: x indexes columns and y indexes rows.\par

\textbf{Task.} Identify the single object instance referred to by the instruction, using its spatial qualifiers. Return a pixel at the center of its visible body. Abstain if the instance is ambiguous or not visible.\par

\textbf{Response.} Return abstain, x, y, confidence, and a brief reason.\par

\end{promptbox}

When association with the frozen reference fails before grasping, the query includes the initial image with the reference highlighted in magenta. The dimensions \texttt{\{W\}} and \texttt{\{H\}}, instruction, phrase, and phase are runtime substitutions.

\begin{promptbox}{Object localization | Reference recovery}
\textbf{Context.} Task instruction: \texttt{\{instruction\}}. Object phrase: \texttt{\{phrase\}}. Execution phase: \texttt{\{phase\}}.\par

\textbf{Images.} Image 1 is the current camera view, \texttt{\{W\}} by \texttt{\{H\}} pixels. Image 2 is the initial view with the reference instance highlighted in magenta. Coordinates in image 1 start at the upper left.\par

\textbf{Task.} Find the same physical instance in image 1, allowing for objects moved by the robot. Similar-looking objects are different instances. Return a pixel at the center of the visible body. Abstain if the reference cannot be identified confidently or is not visible.\par

\textbf{Response.} Return abstain, x, y, confidence, and a brief reason.\par

\end{promptbox}

\Needspace{85mm}
\subsection{Choosing a Whole-Object Mask}
SAM3 can return nested masks at one point. At initial binding, a separate query chooses the complete object rather than an inner surface, lid, merged support, or rival instance. Image 1 provides the full view; subsequent images show the highlighted candidate crops. The response selects one candidate image label or abstains. Reference-conditioned localization instead selects among nested masks using the frozen body's size.

\begin{promptbox}{Mask selection | Whole-object instance}
\textbf{Context.} Task instruction: \texttt{\{instruction\}}. Object phrase: \texttt{\{phrase\}}. Execution phase: \texttt{\{phase\}}.\par

\textbf{Images.} Image 1 shows the full view with a magenta cross on the proposed object. Each subsequent image shows the same cropped region with one candidate mask highlighted.\par

\textbf{Task.} Choose the image highlighting exactly the complete requested object instance. Exclude masks covering only a label, lid, or inner surface; masks merged with a neighboring object or support; and masks of another instance.\par

\textbf{Response.} Return the candidate image label as choice and a brief reason. Return none if no candidate is valid.\par

\end{promptbox}

\subsection{Identity Before Grasping and Transport}
The stack freezes object references before any object moves. Before grasping, it associates a point-prompt mask by majority overlap or a 3\,cm center-distance bound; failed association invokes reference-conditioned localization. A confirmed, unchanged reference can bypass another identity query. Otherwise, the cross-frame prompt below returns \texttt{yes}, \texttt{no}, or \texttt{unsure}.

\begin{promptbox}{Identity verification | Before grasping}
\textbf{Context.} Task instruction: \texttt{\{instruction\}}. Object phrase: \texttt{\{phrase\}}. Execution phase: \texttt{\{phase\}}.\par

\textbf{Images.} Image 1 highlights the instance identified at task start. Image 2 highlights the current candidate.\par

\textbf{Task.} Determine whether both highlighted regions refer to the same physical instance. Account for robot-induced motion and use surrounding unmoved objects as context. Similar appearance alone does not establish identity.\par

\textbf{Response.} Return verdict as yes, no, or unsure, with a brief reason.\par

\end{promptbox}

After grasping, the query compares the frozen reference, the pregrasp confirmation, and the current lifted object. The template accommodates either two or three images: the pregrasp confirmation is included when available.

\begin{promptbox}{Identity verification | After grasping}
\textbf{Context.} Task instruction: \texttt{\{instruction\}}. Object phrase: \texttt{\{phrase\}}. Execution phase: \texttt{\{phase\}}.\par

\textbf{Images.} Image 1 highlights the initial reference. When available, image 2 highlights the pregrasp confirmation. The final image shows the object currently lifted between the fingers.\par

\textbf{Task.} Determine whether the lifted object is the same physical instance as the highlighted reference. Use the earlier arrangement, the grasped location, and any newly empty location as evidence. Abstain from a positive identity claim when the evidence is insufficient.\par

\textbf{Response.} Return verdict as yes, no, or unsure, with a brief reason.\par

\end{promptbox}

Transport requires verified identity, a verified hold, a reached lift, and at least 5\,cm of observed object-base rise. A positive model answer alone is insufficient.

\Needspace{45mm}
\subsection{Structured Responses and Runtime Configuration}
The point-query schema requires \texttt{abstain}, integer \texttt{x}/\texttt{y}, a confidence enum, and a reason. Identity queries require a verdict enum and a reason; mask-choice queries require \texttt{choice} (a candidate image label or \texttt{none}) and a reason. Table~\ref{tab:vlm-response-fields} summarizes the response fields; additional fields are disallowed.

\begin{table}[!ht]
\centering
\small
\caption{Structured outputs for the inference prompts. Every response includes a brief reason.}
\label{tab:vlm-response-fields}
\setlength{\tabcolsep}{5pt}
\renewcommand{\arraystretch}{1.18}
\begin{tabularx}{\linewidth}{@{}>{\raggedright\arraybackslash}p{0.22\linewidth}>{\raggedright\arraybackslash}p{0.32\linewidth}Y@{}}
\toprule
Query & Required fields & Allowed values \\
\midrule
Point localization & \texttt{abstain}, \texttt{x}, \texttt{y}, \texttt{confidence}, \texttt{reason} & Boolean abstention; integer pixel coordinates; confidence: \texttt{low}, \texttt{medium}, or \texttt{high}. \\
Mask selection & \texttt{choice}, \texttt{reason} & A candidate image label, or \texttt{none}. \\
Identity / mask review & \texttt{verdict}, \texttt{reason} & \texttt{yes}, \texttt{no}, or \texttt{unsure}. \\
\bottomrule
\end{tabularx}
\end{table}

We set the model identifier to \texttt{claude-opus-5-5}, effort to \texttt{low}, and the default response limit to 3,000 tokens. It sends images and prompt text as one user message with a JSON-schema output format; it adds no separate system prompt. Errors, refusals, truncation, or budget exhaustion return no usable answer and are handled conservatively by the policy.



\paragraph{Execution.}
Motion-dependent actions use explicit arrival/contact results, and failed transport or release checks invoke bounded recovery. The placement verifier distinguishes \texttt{ready\_to\_release} from \texttt{placed} and measures support geometry, identity, gripper clearance, and two-frame stability. However, this snapshot sets \texttt{enforce=False} and \texttt{corrections=False}: the placement verifier is a shadow monitor, whereas the contact-aware descent gate controls release. At the outer plan level, \texttt{run()} logs a subgoal's return value and starts the next subgoal; it does not require the shadow placement verdict to be \texttt{success}. Consequently, the implementation should not be described as enforcing every placement postcondition at every skill boundary. Official task completion is measured by the evaluator separately from controller logs.

\section{Implementation Details}
\label{app:reproducibility}
The final artifact should include the shared-stack snapshots, source and target adapters, accepted and rejected patch ledger, evaluator implementation, exact model identifiers, environment revisions, and commands reproducing each table. Implementation records are organized by frozen program version and evaluation phase. A version record links the parent and candidate code hashes to the proposed patch, tests, gate decision, and rollout directories. Episode records identify the instruction, perturbation axis, task, seed and initial-state index, program hash, official outcome, termination status, and wall-clock duration. The inference comparison uses the N13 stack and the N13O/N13C driver records; these artifact identifiers name recorded implementations rather than additional learned stacks. A public artifact URL is not yet specified.

\subsection{Grasping-backend Mismatch}
Our simulation experiments use the open-source Contact-GraspNet backend rather than the AnyGrasp\cite{fang2023anygrasp} backend used in the ASPIRE configuration considered for comparison. This mismatch limits the fairness of a direct performance comparison, because differences in grasp proposal quality may affect both individual manipulation outcomes and long-horizon task completion. We speculate that replacing Contact-GraspNet~\citep{sundermeyer2021contact} with AnyGrasp could improve grasp-dependent execution; this effect has not been measured in a matched backend comparison.

\subsection{Adapter and Contract Schemas}
The data-only adapter constructs descriptors from the public runtime instruction. Its fields separate \texttt{objects}, \texttt{supports}, and \texttt{articulated} parts, together with side-word mappings, language templates, anaphora templates, and support prompts. Object and support entries map referring phrases to perception prompts; they do not contain controller code. The adapter takes no benchmark task ID, seed, or simulator state as a policy input. A shared parser produces ordered subgoals, and dispatch checks whether each requested operation is supported before motion begins.

Placement distinguishes support relations from relative placement, including \texttt{on\_support} and \texttt{beside\_support}; later extensions add front/middle reference semantics. The supported vocabulary is versioned with the stack, and unsupported relations are refused rather than silently approximated. Dependencies are expressed by subgoal order and executable readiness checks: completion of one subgoal does not bypass the next skill's preconditions. For example, placing pudding beside a plate in a source instruction and placing it beside a plate after a mug-placement subgoal invoke the same relation-aware implementation with different runtime descriptors.

Motion results expose a status, detail, reason, reached flag, and elapsed \texttt{wall\_s}. Recorded statuses include \texttt{reached}, \texttt{unreachable}, \texttt{stalled}, \texttt{uncertain}, \texttt{invalid\_input}, and \texttt{budget\_exhausted}. A reached motion is distinct from an established placement relation. Motion budgets are measured in control steps; episode limits additionally use wall-clock seconds. In the recorded inference-driver comparison, the common limits are 4,000 simulation steps and 1,500 seconds per episode, including model calls.

\subsection{Regression Gates}
Candidate construction checks exact patch anchors, the declared shared module, forbidden literals in inserted code, Python compilation, and supplied test files. Identifier checks constrain explicit specialization to benchmark tasks or seeds. These automated checks do not establish semantic generality; review additionally examines whether geometric constants follow calibration, measured geometry, or an object-independent construction. Missing tests produce a warning in the inherited builder, while the evaluation protocol requires supporting tests before acceptance.

Mechanism tests exercise the changed behavior and its failure returns. Recorded observations support replay of perception and verification decisions; changes to motion require execution evidence. Parent and candidate are then run from matched source initial states. The gate configuration is fixed at the beginning of each RSI iteration and specifies the evaluation stages, development and selection panels, regression tasks, required improvements, paired-loss limits, and integrity and timeout checks. These settings can evolve across iterations, but are held fixed for all candidate comparisons within a given iteration. The rollout manifest records the corresponding seed-to-state mapping and resource budgets.

Numeric acceptance is followed by inspection of newly successful trajectories and new regression failures. The parent changes only after the checks and review required by the current iteration pass; otherwise the frozen parent is retained. Selection and regression outcomes used by this gate are acceptance evidence, not an independent audit. Any audit panel must be excluded from proposal generation and candidate selection, and reported separately. Target episodes do not supply evidence to the source acceptance gate.

\subsection{Source--Target Combination Generalization}
Source tasks provide manipulation constituents such as grasping, transport, support placement, relative placement, and articulated-object actuation. Target instructions combine these constituents into longer goal sequences. Representative compositions include stove activation followed by pan placement, mug placement followed by pudding placement beside a plate, and bowl containment followed by drawer closure. Object phrases specify the manipulated object and reference; relation types distinguish the required goal; ordered subgoals specify the composition. Reuse is therefore evaluated at the level of shared executable calls rather than instruction-string overlap alone.

The main LIBERO-Pro Long protocol comprises ten tasks with 50 states per task and axis. The separate inference-driver comparison uses five states per task and axis, seeds 41--45, for 100 episodes. These panels have different denominators and are not pooled. Pairing uses task, suite, seed, and recorded initial-state index, with the benchmark revision needed to reconstruct the state. A seed label alone is insufficient to establish identical initial conditions across revisions.

Frozen execution excludes code modification during the target episode. It does not by itself establish that evaluation tasks were never observed during development. The recorded inference comparison reused evaluation rows, and its protocol identifies historical exposure to Scene5. These records must be distinguished from a newly isolated target evaluation when interpreting compositional generalization.

\subsection{Resource Accounting}
Official full-task outcomes are retained for every planned evaluation row, including failures, timeouts, generation refusals, and execution errors. Success is aggregated per task and axis before macro-averaging. Paired comparisons additionally count transitions from parent failure to candidate success and from parent success to candidate failure, since net success can conceal lost capabilities. The reported exact McNemar comparisons use these discordant outcomes. Confidence intervals are not inferred from rounded aggregate rates; they require episode outcomes and a procedure that respects task grouping.

Source improvement and target inference are accounted for separately. Source records distinguish proposal generation, offline tests, candidate rollouts, and rejected candidates. Inference records distinguish initial program generation, runtime decision and perception calls, and robot execution. End-to-end episode time includes generation and execution; model calls inside execution are not added a second time. The figure reports mean duration over all episodes, not only successes. Internal bounded retries remain part of their original episode and consume its budget.

For N13C, the reported amended comparison substitutes five infrastructure-failed rows with their designated reruns; the original records remain available. Smoke episodes and offline generation tests are excluded from the 100 policy outcomes, while their costs remain separate development expenditures. Generated-tool counts should distinguish newly registered operations from revisions to existing ones; invocation traces establish use, whereas matched removal tests are needed to attribute a success gain to a component. No total cost, repeat count, or tool count is inferred where the corresponding ledger is absent.

\section{LIBERO-Pro Long Zero-Shot Evaluation Details}
\label{app:zero-shot-per-task}
\subsection{Task Manifest and Frozen Execution}
The per-task breakdown uses the 10 LIBERO-Pro Long tasks and the \emph{Pos}/\emph{Task} variants listed in ASPIRE's zero-shot appendix~\citep{lu2026aspire}. Descriptive task names below follow that appendix; the artifact must map each name to its exact benchmark task ID, instruction, and initial-state manifest. Each task--axis pair uses seeds 1--50 and a fixed task program. All shared code and adapters are finalized before these episodes; target feedback cannot select alternative programs or patches. The artifact records the source task order, final stack hash, generated program and adapter hashes, benchmark revision, seed-to-state mapping, execution horizon, and recovery allowance.

\subsection{Source-Experience Scaling}
\label{app:transfer-scaling}
\begin{figure}[htbp]
\centering
\includegraphics[width=0.8\linewidth]{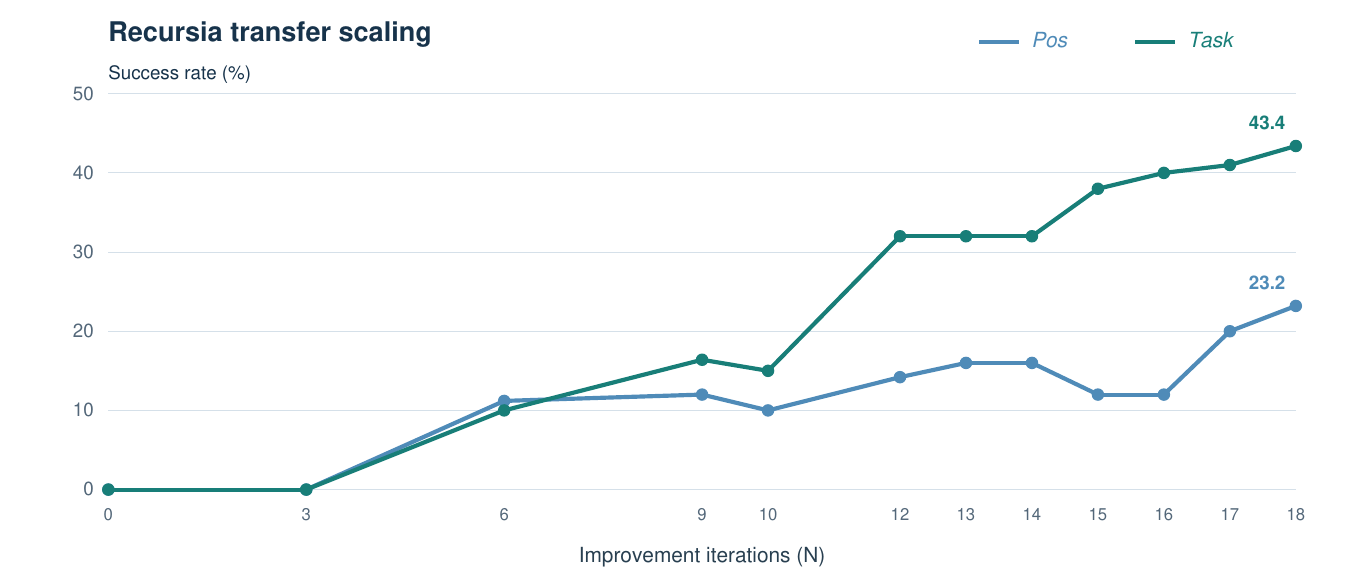}
\caption{\recursia{} transfer scaling with source-task experience. Frozen-stack success rates on \emph{Pos} and \emph{Task} are shown at the supplied iteration checkpoints.}
\label{fig:transfer-scaling}
\end{figure}

The scaling evaluation in Figure~\ref{fig:transfer-scaling} uses one-shot task-program inference at every checkpoint. A task program is generated once and executed over the frozen shared stack without target-side code debugging; its execution may nevertheless use observation feedback. Each plotted value is from a single evaluation run, rather than an average across repeated runs.
Figure~\ref{fig:transfer-scaling} reports the source-experience scaling evaluation for frozen \recursia{} stacks after $N\in\{0,3,6,9,10,12,13,14,15,16,17,18\}$ improvement iterations. Each snapshot is selected using source data only and evaluated without target-driven modification.

\subsection{Aggregation}
With $y_{ias}\in\{0,1\}$ denoting official full-task success on seed $s$, compute
\begin{equation}
\widehat p_{ia}=\frac{1}{50}\sum_{s=1}^{50}y_{ias},\qquad
\widehat p_a=\frac{1}{10}\sum_{i=1}^{10}\widehat p_{ia},\qquad
\widehat p_{\mathrm{Overall}}=\frac{\widehat p_{\mathrm{Pos}}+\widehat p_{\mathrm{Task}}}{2}.
\end{equation}
Our aggregate results are computed from unrounded episode outcomes. Reproduced baseline aggregates retain the source's reported precision. Confidence intervals must account for the grouping of states within tasks, and paired comparisons require matched episode-level outcomes.

\end{document}